%% file: paper_en.tex
\documentclass[journal]{IEEEtran}
\usepackage{amsmath,amsfonts}
\usepackage{algpseudocode}
\usepackage{algorithmicx,algorithm}
\usepackage{hyperref}
\usepackage{array}
\usepackage[caption=false,font=normalsize,labelfont=sf,textfont=sf]{subfig}
\usepackage{caption}
\usepackage{textcomp}
\usepackage{stfloats}
\usepackage{url}
\usepackage{verbatim}
\usepackage{graphicx}
\usepackage{cite}

\usepackage[square,sort&compress,numbers]{natbib}
\usepackage{amsmath}
\usepackage{amsfonts} 
\usepackage{amssymb} 
\usepackage{multirow}
\usepackage{threeparttable}
\usepackage{booktabs}
\usepackage{amsmath}
\usepackage{bbding}
\usepackage{xcolor}
\usepackage{pifont}

\graphicspath{{figs/}}

\begin{document}

\title{RSBuilding: Towards General Remote Sensing Image Building Extraction and Change Detection with Foundation Model
}

\author{
Mingze~Wang$^{1}$,~Lili~Su$^{3}$,~Cilin~Yan$^{1}$,~Sheng~Xu$^{1}$,~Pengcheng~Yuan$^{2}$,~Xiaolong~Jiang$^{2}$~and~Baochang Zhang$^{1, \star}$
\vspace{6pt}
\\
Beihang University$^1$, Xiaohongshu Inc.$^2$, Beijing University of Posts and Telecommunications$^3$
}

\maketitle

\input{secs/01-abs.tex}
\input{secs/02-introduction-v1.tex}
\input{secs/03-related_works-v1.tex}

\input{secs/04-method-v1.tex}
\input{secs/05-experiments-v1.tex}

\input{secs/06-conclusion.tex}

\ifCLASSOPTIONcaptionsoff
  \newpage
\fi

\bibliographystyle{IEEEtran}

\bibliography{IEEEabrv,myreferences}

\end{document}

%% file: secs/01-abs.tex
\begin{abstract}
Buildings constitute not only a significant proportion of man-made structures but also serve as a crucial component of geographic information databases, closely linked to human activities. 
The intelligent interpretation of buildings plays a significant role in urban planning and management, macroeconomic analysis, population dynamics, \textit{etc}.
Remote sensing image building interpretation primarily encompasses building extraction and change detection. However, current methodologies often treat these two tasks as separate entities, thereby failing to leverage shared knowledge. Moreover, the complexity and diversity of remote sensing image scenes pose additional challenges, as most algorithms are designed to model individual small datasets, thus lacking cross-scene generalization.
In this paper, we propose a comprehensive remote sensing image building understanding model, termed RSBuilding, developed from the perspective of the foundation model. RSBuilding is designed to enhance cross-scene generalization and task universality. Specifically, we extract image features based on the prior knowledge of the foundation model and devise a multi-level feature sampler to augment scale information. To unify task representation and integrate image spatiotemporal clues, we introduce a cross-attention decoder with task prompts.
Addressing the current shortage of datasets that incorporate annotations for both tasks, we have developed a federated training strategy to facilitate smooth model convergence even when supervision for some tasks is missing, thereby bolstering the complementarity of different tasks.
Our model was trained on a dataset comprising up to 245,000 images and validated on multiple building extraction and change detection datasets. The experimental results substantiate that RSBuilding can concurrently handle two structurally distinct tasks and exhibits robust zero-shot generalization capabilities. The code will be made available for open-source access at \url{https://github.com/Meize0729/RSBuilding}.
\end{abstract}


\IEEEpeerreviewmaketitle

%% file: secs/02-Introduction-v1.tex
\section{Introduction}

\IEEEPARstart{B}uildings serve as the primary components of man-made surface features, embodying a crucial aspect of specialized geographic information systems that are inherently associated with human activities \cite{bx_stt}. The advent of remote sensing technology has significantly streamlined the procedure of mapping building distribution across broad spatial and temporal scales. Techniques for building extraction and change detection play a fundamental role in interpreting building-related data from remote sensing imagery. These methods offer an optimal and efficient pathway for urban planning and management, as well as macroeconomic analysis~\cite{cd_ban, cd_bit, bx_stt, cd_ttp, rsprompter}.

\begin{figure}[t]
\centering
\resizebox{\linewidth}{!}{
\includegraphics[width=\linewidth]{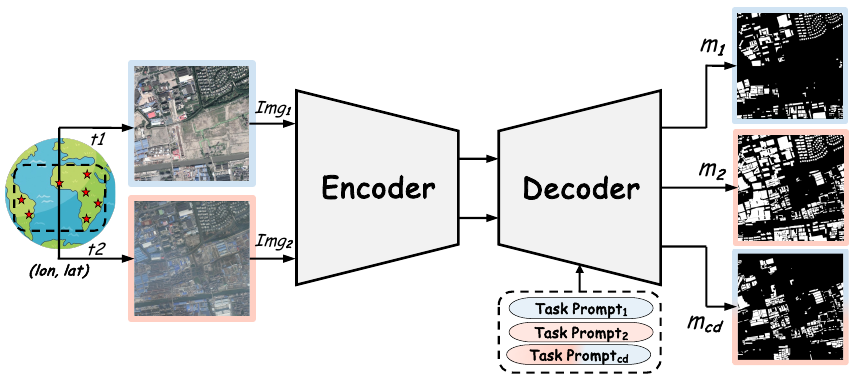}
}
\caption{
The proposed foundation model for remote sensing image building understanding. It possesses the capability to simultaneously output building segmentation masks and change masks across different spatial and temporal points, demonstrating strong generalization ability.
}
\label{fig:teaser}
\end{figure}

Owing to significant advancements in computational capabilities and model development, deep learning technology has progressively become the predominant approach for extracting building-related information from remote sensing imagery \cite{bx_stt, cd_stanet}. Specifically, deep learning-based technologies for building footprint extraction and change detection offer the capacity to interpret single-temporal and multi-temporal remote sensing images, respectively. The former concentrates on the delineation, distribution, and area of buildings, while the latter underscores the construction and demolition of buildings from a diverse range of spatiotemporal remote sensing data \cite{asokan2019change}. Importantly, both tasks converge on the same research subject: the building.

Currently, most methods are specifically designed for individual tasks. Techniques for building extraction strive to overcome challenges such as low segmentation accuracy, blurred edges, small targets, and varying image scales, \textit{etc}. These techniques employ structures such as dual-stream networks, multi-region attention, boundary optimization, and vectorization to augment segmentation results \cite{luo2021deep, bx_stt, bx_buildformer, ghanea2016building}. The capacity of the Transformer \cite{transformer} to model long-range dependencies has further catalyzed the evolution of building extraction algorithms, thereby offering more generalized scene adaptability.
STT \cite{bx_stt} establishes correlations between buildings with minimal cost by leveraging a sparse attention mechanism in both space and channels, thereby achieving superior segmentation performance. BuildFormer \cite{bx_buildformer} tackles the issue of spatial details loss by ingeniously designing a spatial-detailed context path for encoding rich spatial details, alongside a global context path tailored to capturing global dependencies.

Change detection primarily concentrates on the changes of the buildings.
This process not only grapples with the inherent challenges of extracting buildings from single-temporal remote sensing images but also emphasizes the fusion of features across multi-temporal images. Early research predominantly centered around the analysis of differences in input or output, eventually evolving towards feature extraction for the construction of classifiers. However, these methodologies proved insufficient in addressing the complexities of remote-sensing image scenes \cite{shafique2022deep, bai2023deep}.
Contemporary techniques aim at multi-temporal feature fusion, which can be divided into pre-fusion and post-fusion \cite{cd_bit, cd_ttp}. Pre-fusion incorporates temporal information during the feature extraction phase in the backbone, while post-fusion typically gathers temporal information after the spatial features of each image have been extracted. Both strategies harbor their unique advantages and have demonstrated remarkable results. With the emergence and growing popularity of the Transformer model, BiT \cite{cd_bit} has ventured into the utilization of the attention mechanism for the fusion of bi-temporal features. ChangeFormer \cite{cd_changeformer} has established a spatiotemporal correlation by constructing a pure attention feature extraction network. Both have achieved state-of-the-art performance in their respective periods.

Substantial progress in foundation models has paved the way for a more generalized interpretation of remote-sensing images. Within the domain of computer vision, CLIP \cite{clip} has been introduced to align the representational space of images and text by utilizing Noise Contrastive Estimation (NCE) loss. Conversely, SAM \cite{sam} employs a training-annotation loop data engine to conduct supervised training on extensive annotations (11 million images, 1 billion masks), culminating in what is currently the most comprehensive segmentation model. RSPrompter \cite{rsprompter}, a model grounded in prompt learning and built upon the SAM \cite{sam} foundation model, can automatically extract instance-level building masks with minimal learning cost. The BAN \cite{cd_ban} integrates the knowledge of the foundation model into the change detection task through a dual-temporal adapter network. TTP \cite{cd_ttp} introduces an innovative spatiotemporal activation gate and a multi-scale enhancer, thereby enabling the visual foundation model to leverage multi-temporal information and excel in change detection tasks.

Despite significant advancements in the interpretation of buildings from remote sensing imagery, contemporary methodologies predominantly segregate the tasks of building extraction and change detection \cite{cd_bit, bx_buildformer, chen2023continuous_cd}. These approaches overlook the potential benefits of leveraging shared knowledge to enhance performance \cite{chen2023ovarnet}. Furthermore, the intricate scenes and diverse resolutions inherent in remote-sensing images present substantial challenges \cite{chen2023continuous, chen2022resolution}. Existing algorithms concentrate on modeling individual datasets in isolation and configuring hyperparameters independently, thereby constraining their generalizability across multiple scenes and scales \cite{cd_ban, cd_changeformer}. While there have been developments in algorithms based on foundation models, these methods primarily depend on the general knowledge derived from natural image interpretation tasks. They are fine-tuned and transferred on individual datasets and are far from practically applying to the task of interpreting buildings from remote sensing images.

In this paper, we present a general model for remote sensing image building understanding, underpinned by the foundation model. This methodology liberates researchers from the limitations of meticulously crafted subbranches or submodules, thereby enhancing the universality of tasks and generalization across different scenes during interpretation. Our objective is to integrate the tasks of building extraction and change detection within a single framework, thereby investigating their mutual benefits, as illustrated in Fig. \ref{fig:teaser}.
To achieve this, we introduce RSBuilding, a foundation model designed for both building extraction and change detection. Specifically, we introduce a robust backbone for remote sensing imagery based on the Transformer encoder, which abstracts features from dual images, and design a streamlined multi-level feature sampler to augment scale information.
To unify the representations of different tasks and amalgamate associated dual-image features, we propose a cross-attention decoder that operates based on task prompts. Furthermore, to address the existing scarcity of datasets that provide comprehensive annotations for both tasks, we develop a federated training strategy. This strategy facilitates the smooth convergence of the model even in the absence of supervised information for certain tasks, thereby enhancing the complementarity between the two individual interpretation tasks.

\vspace{4pt}
The primary contributions of this paper can be summarized as follows:

i) We propose a general foundation model for building interpretation from remote sensing images, integrating building extraction and change detection tasks within a unified framework, thereby highlighting and leveraging the synergistic relationship between these two tasks.

ii) We introduce a streamlined multi-level feature sampler to construct a feature pyramid, enriching the multi-scale features of buildings. We also propose a cross-attention decoder based on task prompts to aggregate task-specific features and generate the corresponding mask. Additionally, we develop a federated training strategy to address the issue of missing part annotations in multi-task learning.

iii) We have amassed and structured a building dataset consisting of approximately 245,000 images for multi-task training. We validated our pre-trained foundation model on several key datasets, including WHU and INRIA for building extraction, LEVIR-CD and S2Looking for change detection, and BANDON for both tasks. Experimental results indicate that our proposed method is capable of handling two structurally diverse tasks simultaneously and demonstrates robust generalization capabilities across a variety of application scenes.

\vspace{4pt}
The remainder of this paper is organized as follows: Sec. II presents a comprehensive review of the relevant literature. In Sec. III, we delve into the specifics of the structure and training strategy of our proposed RSBuilding. Sec. IV introduces the datasets and the evaluation protocols and provides an in-depth analysis of both quantitative and qualitative results. This section further includes ablation studies, supplemented by a thorough discussion and an acknowledgment of limitations. Lastly, Sec. V concludes the paper by encapsulating the key findings and insights.

%% file: secs/03-related_works-v1.tex
\section{Related Works}

\subsection{Building Extraction}

Remote sensing image building extraction involves pixel-level classification to generate binary masks indicating the presence or absence of buildings, which can be conceptualized as a semantic segmentation problem \cite{luo2021deep}. As one of the fundamental tasks in remote sensing image interpretation, it has broad applications in urban planning, map services, and disaster management.
Historically, building extraction methods primarily employed traditional machine learning techniques such as contour extraction, edge gradients, and gray-level co-occurrence matrices \cite{bx_other1,bx_other2,bx_other3,bx_other4,bx_other5,bx_other6,bx_other7}, \textit{etc.}  However, these methods often struggled to address challenges related to varying scale ranges and shapes, and were also characterized by complex procedures with numerous hyperparameters \cite{ghanea2016building}.

Deep learning presents a more effective approach to these challenges, yielding more accurate segmentation results. Early deep learning-based semantic segmentation methods predominantly utilized the Fully Convolutional Network (FCN) \cite{fcn} architecture, which employs full convolution operations to extract high-level semantic features and deconvolution operations to restore the feature to the original image size. For instance, MA-FCN \cite{bx_mafcn} proposed a multi-scale fusion-based fully convolutional network for building extraction, demonstrating robustness across scales. Subsequently, encoder-decoder structures emerged as the mainstream solution for segmentation tasks. UNet \cite{bx_unet} architecture, which first showed excellent performance in medical image segmentation through its fully convolutional encoder-decoder architecture and skip connections between feature maps of different scales, was adapted for various fields, including building extraction.
However, the limited receptive field of convolution operators remained a significant constraint in obtaining global semantic information. The advent of the Transformer structure led to the development of segmentation networks with attention structures, such as SegFormer \cite{segformer}, which showed remarkable performance. For example, STT \cite{bx_stt} proposed a convolution-transformer hybrid structure with a sparse attention mechanism, striking a balance between accuracy and efficiency. BuildFormer \cite{bx_buildformer} used a spatial context path to encode rich spatial details and a global context path to capture global dependencies. 

Despite these Transformer-based methods achieving impressive results on specific datasets, their generalizability across datasets and scenes remains limited, and they struggle to maintain stable performance in complex and diverse remote sensing scenarios. In this paper, we aim to enhance the model’s general and generalizable capabilities in understanding buildings through the federated training of building segmentation and building change detection tasks.

\subsection{Change Detection}

Change detection, a critical component in the analysis of multi-temporal remote sensing images, is designed to identify variations within specific regions or categories \cite{cd_stanet, cd_bit}. These variations, which may present as either an expansion or contraction of geographical features, yield dynamic insights into human activities and regional economic shifts. Currently, the primary focus of change detection tasks is the monitoring of building alterations due to their significant correlation with human life and economic activities \cite{cd_ban}.

Historically, change detection methodologies were predominantly centered around feature engineering. These methods entailed the construction of multi-temporal features, followed by pixel or region-based classification. However, these approaches were hindered by several limitations, including limited parameter capacity, inadequate adaptability, and restricted generalization capabilities \cite{cd_ban, cd_ttp, wen2021change}.

Recent advancements in deep learning have precipitated a shift towards network-based change detection methods. Leveraging the capabilities of neural networks, these methods are capable of extracting robust image features and performing spatiotemporal fusion across various feature extraction stages effectively \cite{bai2023deep, feng2023change}. 
Early deep learning-based algorithms perceived change detection as a dense prediction task, with the majority exploring different stages of bi-temporal feature fusion to generate the final change segmentation map. 
For instance, Daudt \textit{et al}. \cite{cd_fc} proposed three typical change detection models: FC-EF, FC-Siam-Conc, and FC-Siam-Diff, which generate high-precision change masks by fusing features from different periods. STANet \cite{cd_stanet} introduced an innovative siamese-based spatial-temporal attention neural network, incorporating a self-attention mechanism to effectively model spatial-temporal relationships. This addresses the challenge of capturing spatial-temporal dependencies at various scales.

Recently, the advent of Transformer architectures has introduced new possibilities for the integration of spatiotemporal features in change detection tasks. BiT \cite{cd_bit}, combining Convolutional Neural Network (CNN) and Transformers, establishes an efficient paradigm with a minimal parameter footprint for the fusion of multi-temporal features. ChangeFormer \cite{cd_changeformer}, a model that exclusively relies on Transformer-based siamese networks and fully connected segmentation heads, demonstrates exceptional performance. Changer \cite{cd_changer} verifies the efficacy of feature interactions in change detection by capitalizing on the contextual information derived from paired temporal features.

In this paper, we present a Transformer-based network specifically designed for enhancing building comprehension. By employing federated learning and investigating general knowledge, we aim to augment the precision of building change detection and expand the application scenarios of the proposed model.

\subsection{Foundation Model}

The advent and evolution of foundation models have catalyzed the transformation of artificial intelligence, facilitating the implementation and advancement of more generalized and human-like methodologies across a multitude of application contexts, thereby yielding substantial commercial values. Foundation models, \textit{a.k.a}, base models, predominantly comprise deep learning models characterized by an extensive number of parameters. Owing to their self-supervised or semi-supervised training on voluminous data, these models are capable of quickly adapting to a diverse range of downstream tasks, demonstrating remarkable zero-shot inference capabilities \cite{radford2021learning, jia2021scaling, kirillov2023segment}.

The early iterations of foundation models primarily consisted of pre-trained language models, such as Google’s BERT \cite{devlin2018bert} and OpenAI’s GPT \cite{radford2018improving, radford2019language, brown2020language, openai2023gpt4} series, facilitated by the availability of a vast corpus. These large language models can be rapidly deployed to specific scenarios through fine-tuning, in-context learning, and other techniques \cite{ding2023parameter, salewski2024context}. The introduction of the language dialogue model, ChatGPT \cite{ouyang2022training}, and the multimodal understanding model, GPT-4 \cite{openai2023gpt4}, has effectively initiated research into more versatile foundation models.

Subsequently, a variety of visual foundation models and multimodal foundation models have emerged, including Stable Diffusion \cite{stable_diffusion}, Flamingo \cite{flamingo}, SAM \cite{sam}, \textit{etc}. MiniGPT4 \cite{minigpt4}, by integrating an image encoder with a large language model, can adapt to a wide range of multimodal visual understanding tasks, demonstrating exceptional performance. SAM \cite{sam} leveraging the proposed data annotation engine, trains on up to 1 billion masks, thereby exhibiting comprehensive segmentation capabilities.

In the realm of remote sensing, the majority of contemporary methods adhere to the paradigm of fine-tuning models on downstream tasks by extracting high-level semantic knowledge from remote sensing images or geospatial spaces. For instance, RingMo \cite{sun2022ringmo} acquires a general representation of remote sensing images through the unsupervised image reconstruction training strategy of masked image modeling (MIM) on 2 million remote sensing images. A handful of methods have achieved significant performance improvements with minimal learning costs by transferring the general knowledge of natural image foundation models to the field of remote sensing. RSPrompter \cite{rsprompter} introduces an innovative prompt learning method that incorporates the segmentation knowledge of the SAM foundation model, exhibiting remarkable performance on a wide range of data from different remote sensors. We have developed a general building understanding model and enhanced the model’s generalization capability by amassing a substantial quantity of remote sensing building training data across various resolutions and geographical environments in this paper.

\subsection{Summary}

Current interpretations of buildings in remote sensing imagery are in their infancy, primarily focusing on specific methodologies, datasets, and application scenarios. These approaches often treat the tasks of building extraction and change detection as separate entities. In building extraction, the emphasis is primarily on the exploration of multi-scale building features, precise segmentation of building edges, and refinement of occluded buildings \cite{aleissaee2023transformers}. Conversely, change detection revolves around the extraction of more refined features, efficient temporal feature fusion, and the challenge of scarce positive change samples \cite{shi2020change, li2023lightweight}.

Fundamentally, both building extraction and change detection should align in the low-level semantic space, given that their research subjects are identical - buildings. The divergence lies in the focus: the former concentrates on the existence of buildings, while the latter highlights the changes over time. Our objective is to investigate the complementary between these two tasks within a unified framework, advancing from the perspective of the foundation model, and thereby liberating ourselves from the complexities of subbranch and submodule design. Specifically, we adhere to the widely successful pathway by employing a cumbersome encoder and a lightweight decoder. Additionally, we amass and restructure a substantial volume of annotated building data in various forms, and conduct comprehensive training on these data. This enables the cross-scene and cross-scale understanding of buildings within a unified framework. Our research will provide valuable insights into the evolution of foundation visual models in remote sensing.

%% file: secs/04-method-v1.tex
\section{Methodology}

In this section, we will present an exhaustive introduction to RSBuilding, a general visual foundation model specifically designed for the interpretation of buildings in remote sensing imagery. The discussion will encompass a comprehensive overview of the model, an in-depth explanation of each constituent component, and an examination of the implemented federated training strategy.

\begin{figure*}[t]
\centering
\resizebox{\linewidth}{!}{
\includegraphics[width=\linewidth]{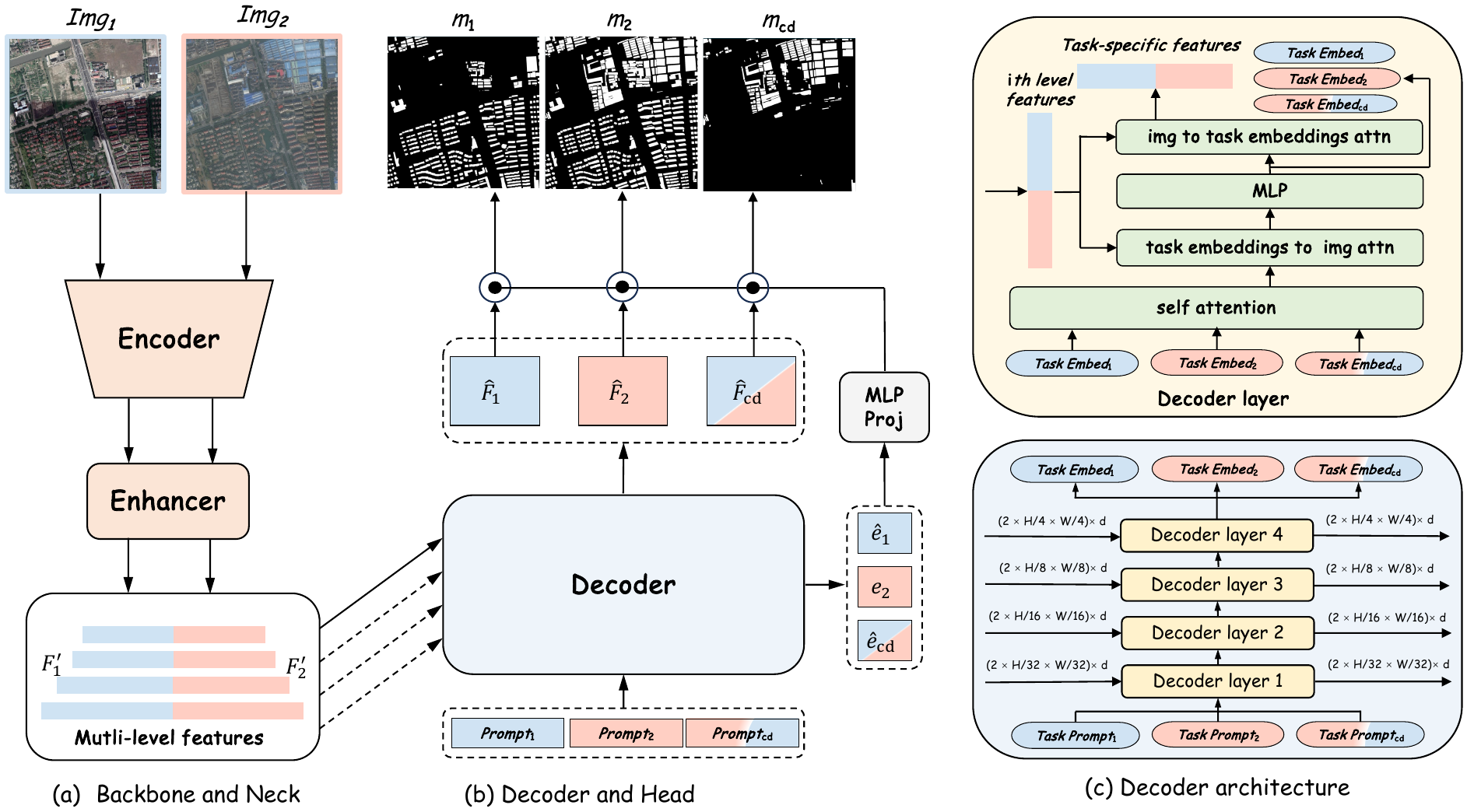}
}
\caption{
The overview structure of the RSBuilding, a model for processing dual-temporal images to obtain individual building masks and building changes. The model primarily consists of four parts: an encoder for extracting dual-temporal robust features, an enhancer for enriching multi-scale information, a decoder for conducting spatiotemporal information interaction and querying corresponding semantic mask features based on task clues, and a segmentation head for generating the final segmentation results by leveraging Einstein summation.
}
\label{fig:main}
\end{figure*}

\subsection{Problem Setting}

Consider a training dataset, denoted as $\mathcal{D}_{\text{train}} = \{(x_1, y_1), \dots, (x_N, y_N)\}$. Each $x_i$ represents a pair of images, specifically $x_i = \{ \mathcal{I}_i^1, \mathcal{I}_i^2 \in \mathbb{R}^{H \times W \times 3} \}$. Correspondingly, $y_i = \{m_i^1, m_i^2, m_i^\text{cd}\}$ signifies the ground-truth annotations associated with $x_i$. Here, $m_i^1, m_i^2 \in \mathbb{R}^{H \times W}$ are the building segmentation annotations for $\mathcal{I}_i^1$ and $\mathcal{I}_i^2$, respectively. Additionally, $m_i^\text{cd} \in \mathbb{R}^{H \times W}$ denotes the change mask between $\mathcal{I}_i^1$ and $\mathcal{I}_i^2$.
It's worth noting that the validity of the building change segmentation mask $m_i^\text{cd}$ hinges on the alignment of geographical spaces $\mathcal{I}_i^1$ and $\mathcal{I}_i^2$.

Our goal is to design and train a foundation model capable of processing any pair of images from a given test set ($x_k \sim \mathcal{D}_{\text{test}}$), concurrently generating the segmentation masks for buildings and the mask indicating changes of buildings over different periods,
\begin{align*}
    \{\hat{m}_k^1, \hat{m}_k^2, \hat{m}_k^\text{cd} \} = \Phi_{\text{head}} \circ \Phi_{\text{dec}} \circ 
 \Phi_{\text{enhancer}} \circ \Phi_{\text{enc}} (x_k)
\end{align*}
where a pair of images undergoes a sequential processing pipeline, comprising a robust encoder, a multi-level feature enhancer and decoder, and a streamlined segmentation head, to produce three binary segmentation masks. The overall architecture is illustrated in Fig. \ref{fig:main}.
For the sake of clarity in the ensuing discussion of the proposed model, the subscript $k$ will be omitted.

\subsection{Encoder}

In this section, we introduce the architecture of the encoder embedded within the RSBuilding model. The encoder, a critical component endowed with a substantial quantity of learnable parameters, facilitates the derivation of robust visual features. Our approach adopts a paradigm wherein dual-temporal general building features are initially extracted independently. Subsequently, we query and express the effective features through the decoder. The encoding process is outlined as follows,
\begin{equation}
\begin{aligned}
    \{ F_1^i \} = \Phi_{\text{enc}} (\mathcal{I}_1) \\
    \{ F_2^i \} = \Phi_{\text{enc}} (\mathcal{I}_2)
\end{aligned}
\end{equation}
where $\{ F_1^i \}$ and $\{ F_2^i \}$ are the respective sets of features extracted from $\mathcal{I}_1$ and $\mathcal{I}_2$. The superscript $i$ represents the $i$th level of the feature.
We employ two prevalent foundation architectures based on the Transformer to establish the encoder. Specifically, the Vision Transformer (ViT) \cite{vit} and the Swin Transformer (Swin) \cite{swin}.

\noindent \textbf{ViT based Siamese Encoder}:
ViT \cite{vit} divides the input image into non-overlapping patches of a fixed size. Each patch is then flattened and linearly projected into a fixed-length embedding. The algorithm employs stacked transformer encoders to model the interrelationships among these embeddings, thereby capturing the global semantic information inherent in the image.
In our paper, we leverage the backbone of the pre-trained SAM \cite{sam} foundation segmentation model and introduce variants with varying parameter quantities. Specifically, we present ViT-B with 12 layers, ViT-L with 24 layers, and ViT-H with 32 layers.
The encoder produces a single-scale feature map, \textit{i.e.}, $\{ F_1^i \} = \{ F_1^1 \in \mathbb{R}^{\frac{H}{2^4} \times \frac{W}{2^4} \times d} \}$ where $d$ represents the channel dimension of the feature map.

\noindent \textbf{Swin based Siamese Encoder}:
The Swin Transformer \cite{swin} constrains the computation of attention to a specific window via a sliding window operation. This approach not only incorporates the locality inherent in Convolutional Neural Network (CNN) but also optimizes the use of computational resources. Different from the ViT, the Swin generates multi-scale feature representations through a hierarchically structured design, \textit{i.e.}, $\{ F_1^i \in \mathbb{R}^{\frac{H}{2^{i+1}} \times \frac{W}{2^{i+1}} \times d^i} \}, i \in \{1,2,3,4\}$.

\subsection{Multi-level Feature Enhancer}

To tackle the complexities associated with multi-resolution remote sensing imagery and objects of varying scales, we introduce a feature enhancer to generate semantic features across multiple scales. The inputs are derived from the outputs of various architectural encoders, ultimately resulting in multi-level features of different sizes but with consistent channel dimensions. The process can be expressed with the following formula,
\begin{equation}
\begin{aligned}
    \{ {F^\prime}_1^i \} = \Phi_{\text{enhancer}} (\{ F_1^i \}) \\
\end{aligned}
\end{equation}
where we only present the single-temporal image feature process for simplicity. The set $\{ F_1^i \}$ represents the feature maps derived from the encoder of $\mathcal{I}_1$. $\{ {F^\prime}_1^i \in \mathbb{R}^{\frac{H}{2^{i+1}} \times \frac{W}{2^{i+1}} \times d} \}, i \in \{1,2,3,4 \}$, denotes the multi-level features. 
Specifically, we employ straightforward max-pooling down-sampling layers and transposed convolutional up-sampling layers to generate multi-level feature maps for the ViT encoder. 
For the Swin encoder, we leverage a standard Feature Pyramid Network (FPN) \cite{fpn} layer.

\subsection{Decoder}

To address the challenges of unifying building extraction and change detection within a framework, we have developed an efficient multi-task decoder that leverages the attention mechanism. More specifically, we employ three learnable query embeddings to aggregate semantic features pertinent to different tasks. These three query embeddings, denoted as $e_1$, $e_2$, and $e_{\text{cd}}$, correspond to the building extraction tasks for $\mathcal{I}_1$ and $\mathcal{I}_2$, and the building change detection task between them, respectively. 
These query embeddings interact with the dual-temporal image features via a cross-attention mechanism, thereby modeling the spatiotemporal relationship and deriving task-specific feature representations. The decoder processes the dual-temporal features at various levels through a series of mixed cross-attention layers. The process can be recursively written as follows,
\begin{equation}
\begin{aligned}
E^0 &= \Phi_{\text{Cat}}(e_1, e_2, e_{\text{cd}}) \\
T_1^i &= {F^\prime}_1^i + \text{PE} + \text{LE}_1 \\
T_2^i &= {F^\prime}_2^i + \text{PE} + \text{LE}_2 \\
T^i &= \Phi_{\text{Cat}}(\Phi_{\text{Flatten}}(T_1^i), \Phi_{\text{Flatten}}(T_2^i))\\
E^i &= \Phi_{\text{S-attn}}(E^{i-1}) \\ 
E^i &= \Phi_{\text{C-attn}}(E^i, T^i) \\
E^i &= \Phi_{\text{mlp-proj}}(E^i) \\
T^i &= \Phi_{\text{C-attn}}(T^i, E^i) \\
\end{aligned}
\end{equation}
where $E^0$ represents the task-specific embedding at the $0$th level. The symbol $\Phi_{\text{Cat}}$ denotes the operation of vector concatenation along the sequence dimension. $\text{PE}$ refers to a pre-set learnable positional encoding map. 
The temporal encodings of $\mathcal{I}_1$ and $\mathcal{I}_2$ are represented by $\text{LE}_1$ and $\text{LE}_2$, respectively. The operation $\Phi_{\text{Flatten}}$ signifies the process of flattening a 2-dimensional feature map into a 1-dimensional sequential vector. 
$T^i \in \mathbb{R}^{(2 \times \frac{H}{2^{i+1}} \times \frac{W}{2^{i+1}}) \times d}$ denotes the $i$th level concatenated dual-temporal image features. The standard multi-head self-attention layer is represented by $\Phi_{\text{S-attn}}$, while the multi-head cross-attention layer is denoted by $\Phi_{\text{C-attn}}$. $\Phi_{\text{mlp-proj}}$ refers to a linear projection layer. 

We follow the operation outlined below to obtain task-specific dense feature representations, which are used to interpret the corresponding masks,
\begin{equation}
\begin{aligned}
\hat{F}_1^i, \hat{F}_2^i &= \Phi_{\text{Reshape}}(\Phi_{\text{Split}}(T^i)) \\
\hat{F}_1^i &= \Phi_{\text{Up}}(\hat{F}_1^i) \\
\hat{F}_1 &= \Phi_{\text{CLR}}(\Phi_{\text{Cat}}(\{ \hat{F}_1^i \})) \\
\hat{F}_\text{cd} &= \hat{F}_1 - \hat{F}_2 \\
\end{aligned}
\end{equation}
where $T^i$ is split and reshaped to reestablish the dual-temporal 2-dimensional feature maps, denoted as $\hat{F}_1^i \in \mathbb{R}^{\frac{H}{2^{i+1}} \times \frac{W}{2^{i+1}} \times d}$ and $\hat{F}_2^i \in \mathbb{R}^{\frac{H}{2^{i+1}} \times \frac{W}{2^{i+1}} \times d}$, where $i$ is an element of the set $\{1,2,3,4\}$. 
The operator $\Phi_{\text{Up}}$ signifies the process of upsampling feature maps to a unified scale of $\frac{H}{4} \times \frac{W}{4}$, achieved through bi-linear interpolation. The operator $\Phi_{\text{Cat}}$ is used to represent the concatenation of feature maps along the channel dimension. 
The term $\Phi_{\text{CLR}}$ denotes a Conv-LayerNorm-ReLU layer with a kernel size of $1 \times 1$ to reduce the channel dimension to 64. This is followed by another Conv-LayerNorm-ReLU layer with a kernel size of $3 \times 3$ to integrate the spatial information. 
$\hat{F}_2$ is generated in the same manner as $\hat{F}_1$. $\hat{F}_\text{cd} \in \mathbb{R}^{\frac{H}{4} \times \frac{W}{4} \times 64}$ denotes the changed dense features between the dual-temporal images.

\subsection{Segmentation Head}

To derive the interpretive results for various tasks, we employ task-specific embedding vectors to filter the dense feature, \textit{i.e.}, Einstein summation, illustrated as follows,
\begin{equation}
\begin{aligned}
\hat{E} &= \Phi_{\text{e-proj}}(E^4) \\
\hat{e}_1, & \hat{e}_2, \hat{e}_\text{cd} = \hat{E} \\
\hat{m}_1 &= \Phi_{\text{e-sum}}(\hat{F}_1, \hat{e}_1) \\
\end{aligned}
\end{equation}
where $\Phi_{\text{e-proj}}$ denotes a linear projection layer to diminish the dimension of the final-layer task-specific embedding. The segmentation mask, $\hat{m}_1$, is computed by $\Phi_{\text{e-sum}}$ through linear weighting $\hat{F}_1$ using $\hat{e}_1$. It should be noted that $\hat{m}_2$ and $\hat{m}_\text{cd}$ are derived in the same manner.

\subsection{Training Strategy}

The objective of our study is to address the tasks of building extraction and change detection from the perspective of the foundation model, all within a unified framework, and validate the universality and complementarity of the intrinsic knowledge shared between the two tasks. Ideally, we aspire to possess an extensive collection of bi-temporal data pairs, $\mathcal{I}_1, \mathcal{I}_2$, accompanied by corresponding labels for building segmentation and change, $m_1, m_2, m_\text{cd}$. However, in practical scenarios, only the BANDON \cite{pang2023detecting} and xView2 \cite{xview2} datasets provide such comprehensive annotations. Unfortunately, the volume of the datasets is limited and falls short of the requirements for high-level generalization. To mitigate this limitation, we have curated and structured a substantial quantity of datasets for building extraction and change detection, collectively referred to as the RSBuilding Dataset. It is noteworthy that the data may solely encompass annotation information pertinent to a single task.

We employ binary Cross-Entropy (CE) loss for the supervised training. The overall loss function is as follows,
\begin{equation}
\begin{aligned}
\mathcal{L} = \alpha_{1} \text{CE}(\hat{m}_1, m_1) + \alpha_{2} \text{CE}(\hat{m}_2, m_2) + \\
\alpha_{\text{cd}} \text{CE}(\hat{m}_\text{cd}, m_\text{cd}) + \alpha_{\text{p-cd}} \text{CE}(|\hat{m}_1 - \hat{m}_2|, m_\text{cd}) \\
\end{aligned}
\end{equation}
where $|\hat{m}_1 - \hat{m}_2|$ denotes the pseudo change detection prediction.
To accommodate a diverse need for tasks and training data within a unified training loss framework, we have the following conventions for $\alpha_{1}, \alpha_{2}, \alpha_{\text{cd}}$, and $\alpha_{\text{p-cd}}$:

\vspace{4pt}
\noindent i) For the building extraction task, which includes building segmentation annotations, we create a duplicate of $\mathcal{I}_1$, denoted as $\mathcal{I}_2$, and apply random photometric distortion. The parameters are set as follows: $\alpha_{1}=1$, $\alpha_{2}=1$, $\alpha_{\text{cd}}=0$ (to mitigate the introduction of negative samples), $\alpha_{\text{p-cd}}=0.1$, and $m_\text{cd} = 0$.

\vspace{4pt}
\noindent ii) For the change detection task, which solely includes change masks, the parameters are adjusted to $\alpha_{1}=0$, $\alpha_{2}=0$, $\alpha_{\text{cd}}=1$, $\alpha_{\text{p-cd}}=0.1$ (to minimize the impact of geometric location errors in bi-temporal images).

\vspace{4pt}
\noindent iii) For the change detection task that includes comprehensive annotations, the parameters are $\alpha_{1}=1$, $\alpha_{2}=1$, $\alpha_{\text{cd}}=1$, $\alpha_{\text{p-cd}}=0.1$.

%% file: secs/05-experiments-v1.tex
\section{Experimental Results and Analyses}

\subsection{Experimental Datasets and Settings}
\label{sec:dataset}

In this paper, we present the RSBuilding Dataset, an extensive and large-scale dataset meticulously curated to train foundation models for various building understanding tasks. This dataset includes annotated data specifically designed for tasks such as building extraction and building change detection.
The entire dataset can be primarily divided into three categories. 

\vspace{4pt}
\noindent i) \textbf{Building Segmentation Masks for Single-temporal Images}: WHU Building Dataset \cite{ji2018fully}, Massachusetts Building Dataset \cite{mnih2013machine}, INRIA Building Dataset \cite{maggiori2017can}, Map Challenge \cite{map_challenge}, and the 2018 Open AI Tanzania Building Footprint Segmentation Challenge \cite{openai_challenge}.

\vspace{4pt}
\noindent ii) \textbf{Building Change Masks for Dual-temporal Images}: EGY-BCD \cite{holail2023afde}, MSBC \cite{9791854}, MSOSCD \cite{9791854}, LEVIR-CD \cite{chen2020spatial}, S2Looking \cite{shen2021s2looking}, and WHU-CD \cite{ji2018fully}.

\vspace{4pt}
\noindent iii) \textbf{Building Segmentation and Change Masks for Dual-temporal Images}: BANDON \cite{pang2023detecting} and xView2 \cite{xview2}.

\vspace{4pt}
An overview of the dataset used for training and testing is provided in Tab. \ref{tab:dataset_tab}. All the data are copped to a unified size of $512 \times 512$. We have selectively gathered partial data from both the Map Challenge and Tanzania Challenge datasets. 
To demonstrate the efficacy of the proposed method, we conducted exhaustive experimental validation using two publicly available building extraction datasets (WHU, INRIA), two change detection datasets (LEVIR-CD, S2Looking), and a dual-label dataset, BANDON. It is imperative to note that the test data relevant to performance validation has been excluded from the RSBuilidng training dataset. The following provides detailed information about these five datasets:

\begin{table}[!thbp]
\centering
\caption{A summary of dataset statistics.
}
\label{tab:dataset_tab}
\resizebox{\linewidth}{!}{
\begin{tabular}{l | *{4}{c}}
\toprule
Dataset & Train & Eval. & Task & Images
\\
\midrule 
WHU \cite{ji2018fully}                     & \checkmark & \checkmark  & Seg & 8k \\
Massachusetts\cite{mnih2013machine}        & \checkmark & \ding{55}   & Seg & 2k \\
INRIA \cite{maggiori2017can}               & \checkmark & \checkmark  & Seg & 12k \\
Map Challenge \cite{map_challenge}        & \checkmark & \ding{55}  & Seg & 90k \\
Tanzania Challenge \cite{openai_challenge}& \checkmark & \ding{55}  & Seg & 6k \\
EGY-BCD \cite{holail2023afde}              & \checkmark & \ding{55}   & CD & 6k \\
MSBC \cite{9791854}                        & \checkmark & \ding{55}   & CD & 4k \\
MSOSCD \cite{9791854}                      & \checkmark & \ding{55}   & CD & 5k \\
LEVIR-CD \cite{chen2020spatial}            & \checkmark & \checkmark   & CD & 3k \\
S2Looking \cite{shen2021s2looking}         & \checkmark & \checkmark   & CD & 20k \\
WHU-CD \cite{ji2018fully}                  & \checkmark & \ding{55}    & CD & 2k \\
BANDON \cite{pang2023detecting}            & \checkmark & \checkmark   & Seg \& CD & 47k \\
xView2 \cite{xview2}                       & \checkmark & \ding{55}    & Seg \& CD & 40k \\
\bottomrule
\end{tabular}
}
\end{table}

\vspace{3pt}
\noindent \textbf{WHU} \cite{ji2018fully}: 
The WHU building dataset is composed of two sub-datasets: one for satellite images and another for aerial images. Our study utilizes the aerial image sub-dataset, which encompasses 8,189 images. These images are further divided into 4,736 for training, 1,036 for validation, and 2,416 for testing. Each image has a spatial resolution of 0.3m. The entire aerial image sub-dataset includes approximately 22,000 buildings, spanning an area exceeding 450 square kilometers.

\noindent \textbf{INRIA} \cite{maggiori2017can}: 
The INRIA building dataset comprises 360 images from five distinct cities: Austin, Chicago, Kitsap, Tyrol, and Vienna. Following the official recommendation, we selected the first to fifth images from each city for validation, with the remaining images allocated for training. The original $5000 \times 5000$ pixel images were cropped to $512 \times 512$ pixels for both training and validation purposes.

\noindent \textbf{LEVIR-CD} \cite{chen2020spatial}: 
The LEVIR-CD dataset includes 637 pairs of bi-temporal remote sensing images, along with over 31,333 instances of change annotation. These images were sourced from Google Earth. Each image pair measures $1024 \times 1024$ pixels, with a resolution of 0.5 meters/pixel. Following the official guidelines, we divided the dataset into 445, 64, and 128 image pairs for training, validation, and testing, respectively.

\noindent \textbf{S2Looking} \cite{shen2021s2looking}: 
The S2Looking dataset contains 5000 image pairs, each measuring $1024 \times 1024$ pixels. These pairs are divided into a $7:1:2$ ratio for training, evaluation, and testing, respectively. The dataset includes over 65,920 instances of change annotation. The images, collected by optical satellites around the world, span a period of one to three years and have a resolution ranging from 0.5 to 0.8 meters/pixel. For training and validation, these images were cropped to a size of $512 \times 512$ pixels.

\noindent \textbf{BANDON} \cite{pang2023detecting}: 
The BANDON dataset is designed to assess the generality and task complementarity of the proposed method. It consists of aerial images with a spatial resolution of 0.6 meters. The images are sourced from various platforms, including Google Earth, Microsoft Virtual Earth, and ArcGIS. The dataset includes geographic images from six representative cities in China: Beijing, Shanghai, Wuhan, Shenzhen, Hong Kong, and Jinan. BANDON comprises 1689 training image pairs, 202 validation image pairs, and 392 testing image pairs, each measuring $2048 \times 2048$ pixels. Each image is accompanied by bi-temporal building segmentation labels and change labels.

\subsection{Evaluation Protocol and Metrics}

We conducted an extensive and comprehensive evaluation of the proposed model using widely accepted metrics in segmentation tasks, including Intersection over Union (IoU), Precision (P), Recall (R), and F1 Score (F1). The definitions of these four evaluation metrics are as follows,
\begin{equation}
\begin{aligned}
\text{IoU} &= \frac{\text{TP}}{\text{TP}+\text{FP}+\text{FN}} \\
\text{P} &= \frac{\text{TP}}{\text{TP}+\text{FP}} \\
\text{R} &= \frac{\text{TP}}{\text{TP}+\text{FN}} \\
\text{F1} &= \frac{2 \times \text{P} \times \text{R}}{\text{P}+\text{R}} \\
\end{aligned}
\end{equation}
where TP, FP, and FN denote true positives, false positives, and false negatives, respectively.

\subsection{Implementation Details} 

Our research is primarily concentrated on the development of a streamlined foundation model architecture, aiming to enhance generalization and versatility in the semantic interpretation of buildings for remote sensing imagery. We approach this problem from the perspective of the foundation model. Unless stated otherwise, our experiments involve multi-task federated training utilizing the amassed RSBuilding dataset, as opposed to being confined to a single dataset.

\vspace{4pt}
\noindent \textbf{Architecture Details}: To accommodate heterogeneous tasks within a unified framework and fulfill the demands of both single-image building extraction and dual-image change detection tasks, we adopt a dual-image input scheme as a compromise. Each image is processed through an encoder and enhancer to extract high-level semantic features. These features are subsequently fused and queried by a decoder to produce three mask maps. For the encoder, we employ either the SA-1B \cite{sam} pre-trained ViT \cite{vit} or the ImageNet 22k \cite{imagenet} pre-trained Swin \cite{swin}. The position encoding of the encoder is adjusted via interpolation to fit a fixed input size of $512 \times 512$.

\vspace{4pt}
\noindent \textbf{Training Details}: During the training phase, we adhere to a fixed input image size of $512 \times 512$ and apply data augmentation techniques such as random cropping, flipping, photometric distortion, mosaic, and temporal exchange. In the testing phase, the images to be validated are cropped to a size of $512 \times 512$, with no additional validation-time augmentation methods applied. We employ binary cross-entropy loss for supervised training and utilize the AdamW optimizer with an initial learning rate of $1e-4$, a weight decay of 0.05, and a Poly scheduler to decay the learning rate. The batch size for training is set at 16. The foundation model underwent 400k training iterations, and the experiment was conducted on 8 NVIDIA A100 GPUs. The algorithm was developed based on the Open-CD~\cite{cd_changer} and MMSegmentation~\cite{mmseg} platform, with Pytorch serving as the underlying deep learning framework.

\input{table/table_bx}

\input{table/table_cd}

\input{table/table_all}

\subsection{Comparison with the State-of-the-Art}

We evaluate the effectiveness of our proposed method, RSBuilding, by comparing it with several state-of-the-art methods for building extraction and change detection. The methods chosen for comparison encompass a wide range of approaches, including: General semantic segmentation methods such as UNet \cite{bx_unet}, Deeplab-v3 \cite{bx_deeplabv3}, and SegFormer \cite{bx_segformer}. Remote sensing image semantic segmentation and building extraction methods, including MA-FCN \cite{bx_mafcn}, DSNet \cite{bx_dsnet}, MSNet \cite{bx_MSNet}, BOMSNet \cite{bx_bomsnet}, LCS \cite{bx_lcs}, Buildformer \cite{bx_buildformer}, BCTNet \cite{bx_bctnet}, and FD-Net \cite{bx_fdnet}. Change detection methods such as FC-Siam-Conc, FC-Siam-Diff \cite{cd_fccdn}, SNUNet-C32 \cite{cd_snunet}, BiT \cite{cd_bit}, Changeformer \cite{cd_changeformer}, TinyCD \cite{cd_tinycd}, HANet \cite{cd_hanet}, and ChangerEx \cite{cd_changer}.
The performance metrics presented for these methods are based on either their officially published results or our re-implementation using PyTorch, adhering closely to the specifications provided in their respective publications.

\subsubsection{Quantitative and Qualitative Comparisons for Building Extraction}

In this section, we will demonstrate the performance of RSBuilding in building extraction tasks using the Swin-B as the backbone model, with quantitative metrics compared as shown in Tab. \ref{tab:bx_tab}. The proposed method performs optimally on both the WHU and INRIA datasets, with an IoU of 92.15\% and 82.68\%, respectively. When compared with other contemporary methods, our approach continues to exhibit significant advantages. This verified the outstanding performance of the proposed method in building extraction tasks.

Accurately representing complex building shapes, semantics, and edges is crucial for effective building extraction. Fig. \ref{fig:bx_fig} visually illustrates the performance of our proposed method compared with other advanced methods. Specifically, 
(a) highlights RSBuilding’s proficiency in segmenting rare-shaped buildings and small-scale targets. 
(b) showcases its precise semantic expression, preventing misclassification of densely arranged containers as buildings. 
(c) and (d) demonstrate its robust discriminative power in distinguishing complex and confusing remote sensing backgrounds. 
(e) exemplifies its global understanding capability, enabling it to handle buildings at different resolutions, including large structures spanning the entire image. 
(f) and (g) underscore its precision edge expression abilities.
Overall, RSBuilding exhibits strong cross-scene generalization, adapting effectively to various complex scenarios and multi-scale transformations.

\subsubsection{Quantitative and Qualitative Comparisons for Change Detection}

The comparative results of change detection tasks between RSBuilding and other methods are delineated in Tab. \ref{tab:cd_tab}. We employ ViT-L as the foundation backbone of RSBuilding. Notably, RSBuilding outperforms other comparative methods on the LEVIR-CD and S2Looking datasets with an IoU of 86.19\% and 52.46\%, respectively. It exhibits robust proficiency in discerning temporal changes within remote sensing imagery.

The effectiveness of building change detection hinges on the ability to discern meaningful change amidst irrelevant variations. Fig. \ref{fig:cd_fig} highlights the exceptional performance of RSBuilding. Specifically, 
(a) demonstrates RSBuilding's capability in managing complex environments and processing degraded images. 
(b) and (c) highlight its prowess in describing changes across various scales of targets. 
(d) demonstrates a more precise extraction of change edges. 
(e) and (f) exhibits an impressive ability to perceive high-level semantic information related to changes. 
This enables more accurate change delineation compared to alternative methods that struggle to define changes and tend to produce more false positives.
Moreover, RSBuilding’s robust generalization is commendable. Even without single-temporal building segmentation mask supervision annotations, it can accurately extract building masks for each corresponding temporal image, thanks to its multi-task federated training approach.

\subsubsection{Quantitative and Qualitative Comparisons for Both Tasks}

To evaluate the performance of both building extraction and change detection tasks, we conducted an extensive evaluation using the BANDON dataset with ViT-L as the backbone, which provides annotations for both tasks. The experimental results indicate that RSBuilding achieves impressive IoU scores of 80.55\% and 58.68\% for the respective tasks, as illustrated in Tab. \ref{tab:all_tab}. 
Significantly, these metrics outperform those achieved by task-specific training methods. The experimental observations not only emphasize the complementary relationship between the two tasks but also underscore the multi-task representation capabilities inherent in the proposed framework.

The segmentation results of the proposed method for two distinct tasks are illustrated in Fig. \ref{fig:all_fig}. We can make the following observations: (a) displays a large building located in the lower-right corner. This building exhibits noticeable differences between time periods A and B, primarily attributed to decorative modifications rather than any construction or demolition activities. Consequently, we argue that these alterations should not be considered as changes. Unlike other methods, RSBuilding delves deeper into the fundamental ``essence” of changes, effectively minimizing irrelevant variations. (b) exemplifies the dual capabilities of our method: building extraction and change detection. Specifically, RSBuilding can accurately delineate edges and represent complex architectural forms. Consequently, the predictions closely correspond with the actual shapes of the buildings, thereby enhancing the precision. (c) presents densely arranged yet non-contiguous buildings. The existence of shadows between building intervals poses a significant challenge in distinguishing individual structures. In this context, RSBuilding not only provides more precise edge predictions but also reduces interference from shadows and other confounding factors during the process of identifying individual buildings.

\subsection{Ablation Study}

In this section, we aim to validate the efficacy of the proposed components and strategies by conducting a series of ablation experiments. Unless otherwise specified, we employ the ViT-L, pre-trained on the SA-1B dataset, as the backbone of the proposed model.

\subsubsection{Effects of Different Encoder Architectures}

The parameter size of the visual foundation model's image encoder can significantly affect the inference speed. An encoder architecture with a robust feature representation capability can yield better performance within a relatively low inference latency. In our study, we conduct experiments utilizing the SA-1B pre-trained ViT and the ImageNet pre-trained Swin Transformer as the encoders respectively. The Swin Transformer exhibits advantages in modeling data with varying scales, while the ViT is more concise, demonstrating a strong inductive bias capability. Specifically, we employ Swin-T, Swin-B, ViT-B, and ViT-L for comparative experiments.

Tab. \ref{tab:backbone_tab} delineates the performance of various encoder architectures in building interpretation. The experimental results indicate that when all models are base versions with similar parameter quantities ($\sim80$M), Swin exhibits certain advantages in modeling buildings and shows better performance on building segmentation. As the model size is further expanded to the large version, there is a corresponding improvement in performance indicators. On the whole, the architectures of Swin and ViT exert a marginal influence on the performance.

\input{table/table_backbone}

\input{table/table_components}

\subsubsection{Effects of Model Components}

To achieve a general model for multi-tasking and to extract more generalized feature representations, we have introduced a multi-scale feature enhancer, a dual-path cross-attention decoder, and a multi-level decoding mechanism into our approach. To validate the performance improvement attributed to these specifically designed sub-modules, we executed a series of ablation experiments. The results are illustrated in Tab. \ref{tab:components}.
Our observations are as follows:
i) Considering the diverse resolution of remote sensing images and the variable scale of building targets, we incorporated a multi-level feature enhancer into the model. The performance enhancements observed in the two tasks substantiate the effectiveness of this design.
ii) Compared to the single-path cross-attention derived from the plain Transformer, the dual-path cross-attention (which mutually constructs attention through preset task-prompt tokens and dual image feature tokens) yielded an IoU increase for both building extraction and change detection tasks. This implies that the bidirectional attention decoder can facilitate a more efficient interaction between image features and query embeddings. 
iii) Given the significance of integrating high-level and low-level semantics in decoding, we separately incorporated multi-level features with task-specific semantic tokens during the decoding process, and unified the resulting semantic features to the same scale for concatenation. The experimental results indicate that this design further contributes to performance enhancements. The ablation studies demonstrate the effectiveness of the components in our model.

\subsubsection{Effects of Federated Training}

We leveraged the amassed data from both building extraction and change detection tasks for the multi-task federated training. To evaluate the impact of each data segment on the model's performance, we employed three different dataset settings: the building extraction dataset (BX), the change detection dataset (CD), and the foundation dataset (BX \& CD), \textit{i.e.}, all data for model pre-training. After pre-training on these data segments, we have individually fine-tuned the pre-trained model on two downstream validation datasets. The experimental results are presented in Tab. \ref{tab:data}.
As discernible from the table, utilizing either individual data segments or the entire dataset can all result in performance enhancements across both tasks.
These insights not only corroborate the complementary nature of the two tasks but also validate the efficacy of the proposed architecture in synergistically handling two heterogeneous tasks and achieving performance enhancements.

\input{table/table_training_data}

\subsubsection{Generalization Ability}

\input{table/table_generalization}

We aim to propose a general and generalized model for the building interpretation. To validate the generalization capability of the proposed method across diverse scenes and datasets, we have conducted experiments on the WHU-EAST ASIA \cite{ji2018fully} and SECOND \cite{SECOND} datasets, following three validation settings: zero-shot generalization, linear probes, and full parameter fine-tuning. 
The results were simultaneously compared with other advanced mainstream methods. Notably, our RSBuilding dataset was not contaminated by these two datasets, \textit{i.e.}, these data were not incorporated into the training data.

The WHU-EAST ASIA dataset \cite{ji2018fully} comprises six contiguous satellite images that cover an area of 860 square kilometers in East Asia, with a spatial resolution of 0.45 meters. This dataset is frequently employed to assess the generalization capability of models across different data sources. We conducted experiments using the second portion of this data, specifically, 3,136 images for training and 904 images for testing.

The SECOND \cite{SECOND} dataset is a multi-category semantic change detection dataset that covers cities such as Hangzhou, Chengdu, and Shanghai. It contains 4,662 pairs of $512 \times 512$ aerial images. Our focus was on changes in the building category, with other categories being treated as background. As only the training set of this data is available, we divided the 2,969 image pairs in the training set into training and testing sets at a ratio of $8:2$.

To thoroughly verify the generalization capability of the pre-trained foundation RSBuilding model, we conducted experiments in three settings: i) Zero-shot, where the foundation model trained on the RSBuilding dataset is directly used for the building understanding tasks; 2) Linear probes, where the majority of the model parameters are frozen and only the segmentation head of the foundation model is fine-tuned; 3) Fine-tuning, where all parameters of the model are further trained following the pre-training and then fine-tuning paradigm.

The experimental results are presented in Tab. \ref{tab:table_generalization}. As can be seen from the table, even within the zero-shot setting, the model's performance achieves a level comparable to other supervised training methods and even surpasses some mainstream methods in the building extraction task. Through linear probing and supervised fine-tuning, the performance is further enhanced, surpassing other methods in all metrics. These findings underscore that the general building interpretation model proposed in this paper possesses robust generalization ability, and can compete with current advanced methods even without any fine-tuning.

\subsection{Discussions}

In this paper, we introduce a general multi-task understanding framework for buildings. Within this framework, we propose a multi-task federated training strategy that accommodates partially missing task labels, thereby validating the complementarity of building extraction and building detection tasks. 
The model is constructed on the widely-used Transformer structure, integrating the fundamental ViT and Swin Transformer structure as the large-scale visual encoder.
During experiments, we verified the performance of both the base and large versions. The results indicated a minimal performance discrepancy between ViT and Swin, with an observed increase in performance correlating with the expansion of the model architecture. Owing to the design's foundation on the prevailing visual model architecture, our model is dynamically scalable and allows for the selection of different foundation model knowledge bases and visual backbones according to the application scenario.

To accommodate the unification of heterogeneous tasks, specifically, the building extraction task involving single image input and the change detection task involving double image inputs, we have duplicated the single building image. While this approach may seem inelegant and results in repeated calculations, it serves as a necessary compromise. Future research could explore the use of the cross-attention mechanism to query varying quantities of image inputs dynamically.

Our ultimate objective is to construct a generally applicable model to serve specific downstream building interpretation tasks. Experimental results affirm the robust generalizability of our method. Depending on the specific downstream application scenarios, researchers have the flexibility to choose from various transfer learning methods, including zero-sample direct application, few-sample linear probing, medium-sample parameter fine-tuning, \textit{etc.}

%% file: table/table_bx.tex
\begin{table*}[!htbp]
\centering
\caption{
Comparison of building extraction performance using various methods on WHU and INRIA test sets.
}
\label{tab:bx_tab}
\resizebox{0.85\linewidth}{!}{
    \begin{tabular}{c|c|cccc|cccc}
    \toprule
    \multirow{2}{*}{Method}& \multirow{2}{*}{Year} & \multicolumn{4}{c|}{WHU (\%)} &\multicolumn{4}{c}{INRIA (\%)}                                           
     \\ 
     & & P & R & F1   & IoU 
     & P & R & F1   & IoU  \\ 
    
    \midrule
    
    UNet \cite{bx_unet} & 2015 
    &94.52 &94.77 &94.52 &89.60 
    &87.21 &86.09 &86.65 &76.44
    \\
    Deeplab-v3 \cite{bx_deeplabv3} & 2017
    &94.80 &93.94 &94.36 &89.33
    &90.30 &87.74 &89.00 &80.18
    \\
    Segformer \cite{bx_segformer} & 2021 
    &94.72 &94.42 &94.57 &89.70
    &88.97 &87.08 &88.01 &78.59
    \\
    \midrule
    HRNet \cite{bx_hrnet} & 2019 
    &91.69 &92.85 &92.27 &85.64
    &86.56 &84.92 &85.73 &75.03 
    \\ 
    MA-FCN \cite{bx_mafcn} & 2019 
    &95.20 &95.10 &95.15 &90.70
    &89.82 &87.58 &88.68 &79.67
    \\ 
    DSNet \cite{bx_dsnet} & 2020 
    &94.05  &94.91 &94.48 &89.54
    &90.32	&88.73 &89.52 &81.02 	
    \\
    MSNet \cite{bx_MSNet} & 2022 
    &94.83 &93.12 &93.96 &89.07
    & - &- &- &-
    \\ 
    BOMSNet \cite{bx_bomsnet} & 2022 
    &95.14 &94.50 &94.80 &90.15
    &87.93 &87.58 &87.75 &78.18
    \\ 
    LCS \cite{bx_lcs} & 2022 
    &95.38 &94.86 &95.12 &90.71   
    &89.58 &86.77 &88.15 &78.82		 
    \\
    BuildFormer \cite{bx_buildformer} & 2022 
    &95.15 &95.14 &95.14 &90.73
    &90.65 &88.78 &89.71 &81.24 
    \\
    BCTNet \cite{bx_bctnet} & 2023  
    &95.47 &95.27 &95.37 &91.15
    &- & -&- &- 
    \\
    FD-Net \cite{bx_fdnet} &2023 
    &95.27 &95.46 &95.36 &91.14 
    &- &- &- &- 
    \\ 
    
    \midrule
    
    RSBuilding (Ours) &-
    & \textbf{95.93} & \textbf{95.82} & \textbf{95.88} & \textbf{92.15} 
    & \textbf{91.40} & \textbf{89.65} & \textbf{90.52} & \textbf{82.68}
    \\ 
    
    \bottomrule
    \end{tabular}
}
\end{table*}

\begin{figure*}[!htpb]
\centering
\resizebox{0.8\linewidth}{!}{
\includegraphics[width=0.8\linewidth]{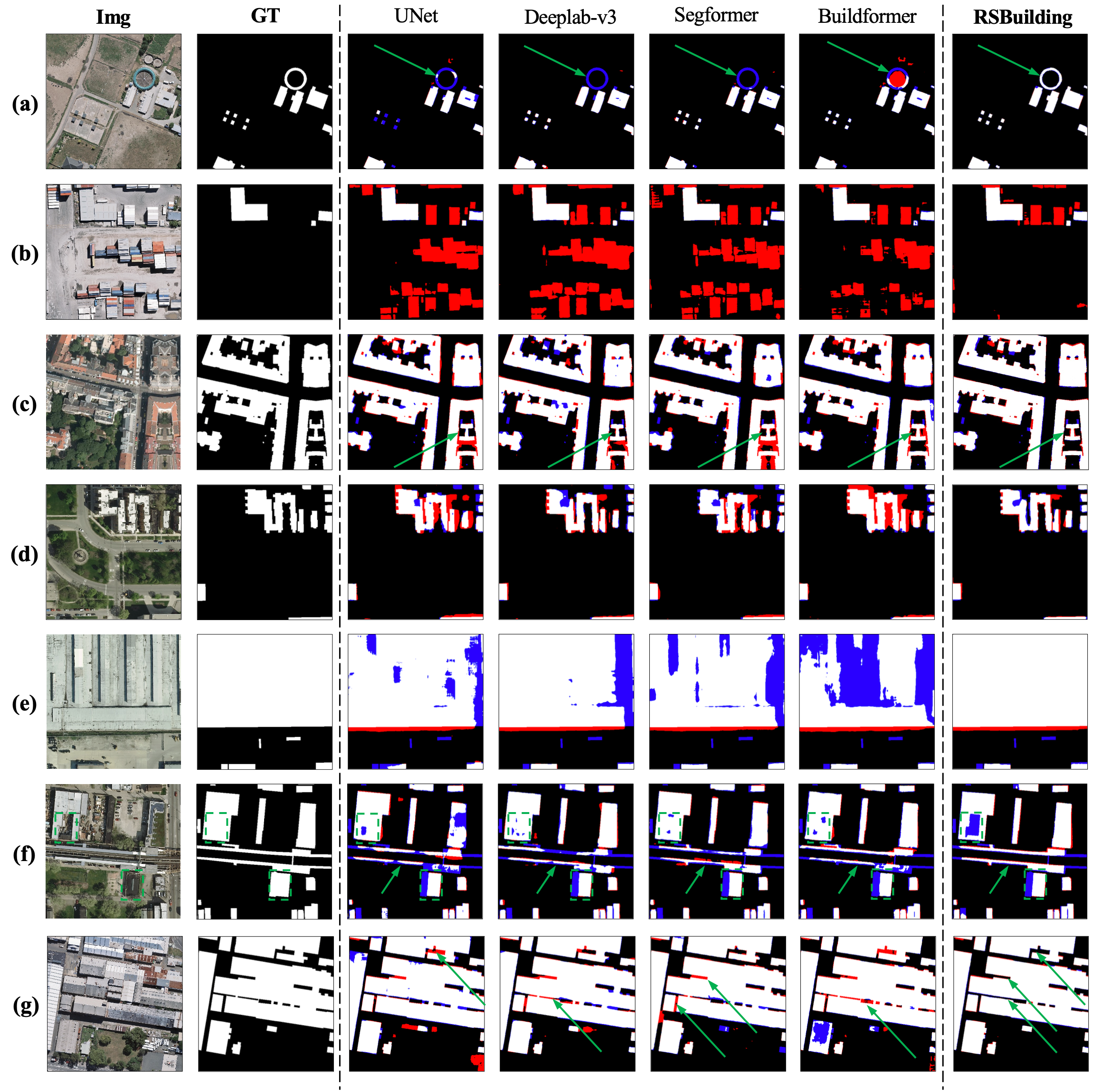}
}
\caption{
Visual comparisons of the proposed method with other state-of-the-art methods for building extraction. \textcolor{red}{Red} means false positives (FP), while \textcolor{blue}{Blue} denotes false negatives (FN). Samples are all from the WHU and INRIA building test sets. 
}
\label{fig:bx_fig}
\end{figure*}

%% file: table/table_cd.tex
\begin{table*}[!htbp]
\centering
\caption{
Comparison of change detection performance using various methods on LEVIR-CD and S2Looking test sets.}
\label{tab:cd_tab}
\resizebox{0.85\linewidth}{!}{
    \begin{tabular}{c|c|cccc|cccc}
    \toprule
    \multirow{2}{*}{Method} & \multirow{2}{*}{Year} & \multicolumn{4}{c|}{LEVIR-CD (\%)} &\multicolumn{4}{c}{S2Looking (\%)} \\
     & & P & R & F1   & IoU 
     & P & R & F1   & IoU  \\ 
    
    \midrule
    
    FC-Siam-Conc\cite{cd_fc} & 2018
    &87.09 &90.10 &88.57 &79.48  
    &\textbf{73.59}	&30.09	&42.72	&27.16
    \\
    FC-Siam-Diff\cite{cd_fc} & 2018
    &88.47 &89.55 &89.00 &80.19  
    &72.39 &34.53 &43.95 &28.16
    \\ 
    SNUNet-C32 \cite{cd_snunet} & 2021 
    &\textbf{93.45} &90.54 &91.97 &85.14 
    &70.53 &58.21 &63.77 &46.81
    \\ 
    BiT \cite{cd_bit} & 2021
    &93.11 &90.21 &91.64 &84.56
    &72.96 &56.38 &63.61 &46.63
    \\ 
    Changeformer \cite{cd_changeformer} & 2022
    &93.03 &89.66 &91.32 &84.02 
    &71.41 &61.42 &66.04 &49.30
    \\
    TinyCD \cite{cd_tinycd} & 2023
    &92.39 &92.84 &90.51 &84.53 
    &71.07 &58.48 &64.16 &47.24
    \\ 
    TinyCDv2-L \cite{cd_tinycd} & 2023
    &92.40 &90.91 &91.64 &84.58
    &72.01 &54.08 &61.77 &44.69
    \\ 
    HANet \cite{cd_hanet} & 2023
    &93.07 &89.42 &91.21 &83.84
    &63.51 &56.36 &59.72 &42.57
    \\ 
    ChangerEx-RS50 \cite{cd_changer} & 2023
    &90.48 &\textbf{93.09} &91.77 &84.79
    &71.96 &61.88 &66.54 &49.86
    \\
    
    \midrule
    
    RSBuilding (Ours) &  -
    &93.39 &        91.80  &\textbf{92.59} &\textbf{86.19} 
    &73.14 &\textbf{64.97} &\textbf{68.81} &\textbf{52.46}
    \\ 
    
    \bottomrule
    \end{tabular}
}
\end{table*}

\begin{figure*}[!htpb]
\centering
\resizebox{\linewidth}{!}{
\includegraphics[width=\linewidth]{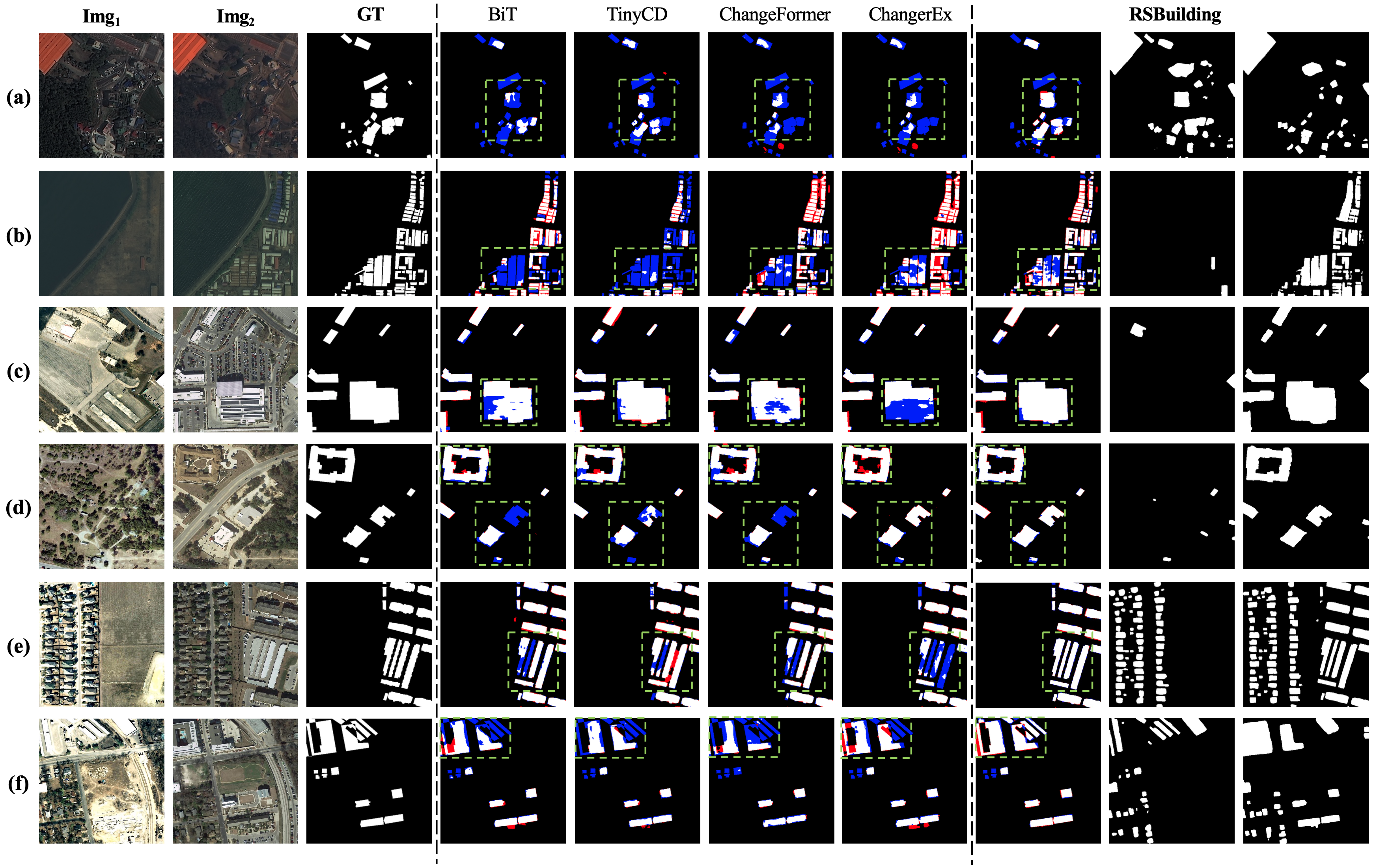}
}
\caption{
Visual comparisons of the proposed method with other state-of-the-art methods for change detection. \textcolor{red}{Red} means false positives (FP), while \textcolor{blue}{Blue} denotes false negatives (FN). Samples are all from the LEVIR-CD and S2Looking test sets. 
The final two columns depict building extraction results from the pre-temporal and post-temporal images.
}
\label{fig:cd_fig}
\end{figure*}

%% file: table/table_all.tex
\begin{table*}[!htbp]
\centering
\caption{Comparison of building extraction and change detection performance using various methods on the BADNON test set.}
\label{tab:all_tab}
\resizebox{0.85\linewidth}{!}{
    \begin{tabular}{c|c|cccc|cccc}
    \toprule
    \multirow{2}{*}{Method}& \multirow{2}{*}{Year} & \multicolumn{4}{c|}{Building Extraction (\%)} &\multicolumn{4}{c}{Change Detection (\%)}                                           
     \\ 
     & & P & R & F1   & IoU 
     & P & R & F1   & IoU  \\ 
    
    \midrule
    
    UNet \cite{bx_unet} & 2015 
    &86.60 &86.93 &86.76 &77.19
    &- &- &- &-
    \\
    Deeplab-v3 \cite{bx_deeplabv3} & 2017
    &86.48 & 88.74 &87.66 &78.53
    &- &- &- &-
    \\
    Segformer \cite{bx_segformer} & 2021 
    &86.81 &87.79 &87.30 &78.05
    &- &- &- &-
    \\
    Buildformer \cite{bx_buildformer} & 2022
    &82.26 &88.22 &87.74 &78.75
    &- &- &- &-
    \\
    \midrule
    
    FC-Siam-Diff \cite{cd_fc} & 2018
    &- &- &- &-
    &72.80 &31.84 &44.31 &28.46
    \\ 
    SNUNet-C32 \cite{cd_snunet} & 2021 
    &- &- &- &-
    &70.43 &56.97 &62.99 &45.97
    \\ 
    BiT \cite{cd_bit} & 2021 
    &- &- &- &-
    &71.16 &65.64 &68.29 &51.85
    \\
    Changeformer \cite{cd_changeformer} & 2022 
    &- &- &- &-
    &74.48 &66.24 &70.12 &53.98
    \\ 
    TinyCD \cite{cd_tinycd} & 2023 
    &- &- &- &-
    &73.01 &62.71 &67.47 &50.91
    \\ 
    ChangerEx \cite{cd_changer} & 2022 
    &- &- &- &-
    &72.69 &68.56 &70.56 &54.52
    \\ 
    
    \midrule
    
    RSBuilding (Ours) &-
    & \textbf{88.82} & \textbf{89.64}  & \textbf{89.23} & \textbf{80.55} 
    & \textbf{75.55} & \textbf{72.44} & \textbf{73.96} & \textbf{58.68}
    \\
    
    \bottomrule
    \end{tabular}
}
\end{table*}

\begin{figure*}[!htpb]
\centering
\resizebox{0.9\linewidth}{!}{
\includegraphics[width=0.9\linewidth]{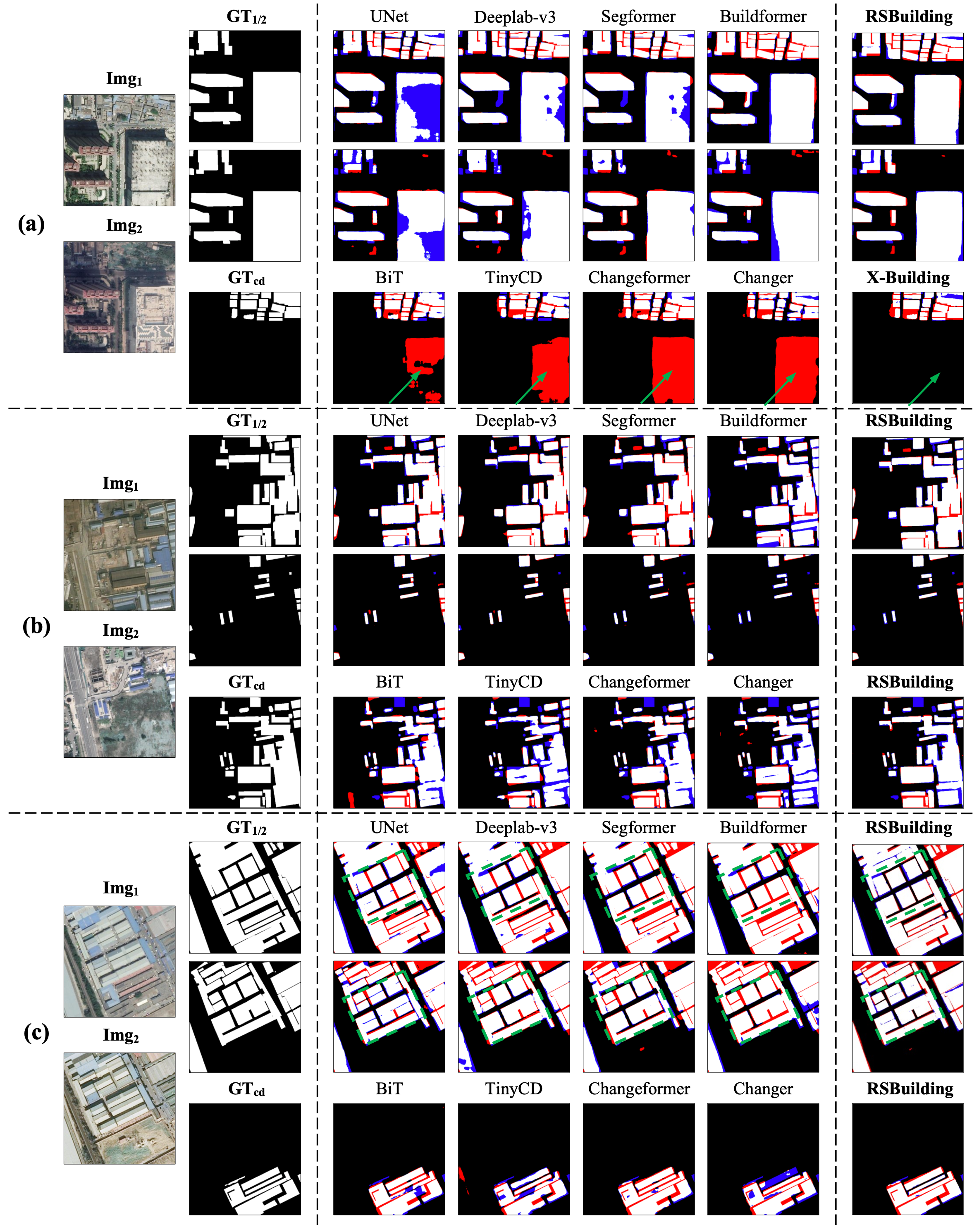}
}
\caption{
Visual comparisons of the proposed method with other state-of-the-art methods for building extraction and change detection. \textcolor{red}{Red} means false positives (FP), while \textcolor{blue}{Blue} denotes false negatives (FN). Samples are all from the BANDON test sets. 
}
\label{fig:all_fig}
\end{figure*}

%% file: table/table_backbone.tex
\begin{table*}[!htbp]
\centering
\caption{The effects of different encoder architectures. Performances are validated on the WHU, INRIA, LEVIR-CD, S2Looking, and BANDON test sets. BANDON-BX means building extraction performance, while BANDON-CD denotes change detection performance. Para. represents the parameter quantity.
}
\label{tab:backbone_tab}
\resizebox{0.99\linewidth}{!}{
\begin{tabular}{c| c |cc|cc|cc|cc|cc|cc}
\toprule
\multirow{2}{*}{Architecture} & \multirow{2}{*}{Para. (M)} & \multicolumn{2}{c|}{WHU} & \multicolumn{2}{c|}{INRIA} & \multicolumn{2}{c|}{LEVIR-CD} & \multicolumn{2}{c|}{S2Looking} & \multicolumn{2}{c|}{BANDON-BX} & \multicolumn{2}{c}{BANDON-CD} \\ 
& & IoU        & F1         & IoU          &F1   & IoU           & F1            & IoU            & F1            & IoU           & F1            & IoU           & F1  \\ 
\midrule
Swin-T        &28            & 91.25       & 95.42      & 81.95        & 90.08       & 85.06         & 91.93         & 50.68          & 67.26         & 79.37          & 88.50         & 57.26         & 72.82         \\
Swin-B &88 & 92.15 & 95.88 & 82.68 & 90.52 & 85.57  & 92.23  & 51.59   & 68.01         & 80.01          & 88.90         & 58.21         & 73.59 
\\
\midrule
ViT-B          &89    & 90.74       & 95.15      & 82.03        & 90.13       & 85.76         & 92.34         & 50.76          & 67.34         & 80.07          & 88.93         & 58.06         & 73.47         \\
ViT-L &308 & 91.36       & 95.42      & 82.59        & 90.46 & 86.19 & 92.59 & 52.46 & 68.81 & 80.55 & 89.23 & 58.68 & 73.96
\\
\bottomrule
\end{tabular}
}
\end{table*}

%% file: table/table_components.tex
\begin{table}[!htbp]
\centering
\caption{The effects of different model component designs (multi-scale feature enhancer, dual-path cross-attention decoder, and multi-level decoding mechanism). Performances are validated on the WHU and INRIA test sets. 
}
\label{tab:components}
\resizebox{0.99\linewidth}{!}{
\begin{tabular}{ccc|cc|cc}
\toprule
\multirow{2}{*}{enhancer} &\multirow{2}{*}{decoder} &\multirow{2}{*}{decoding} & \multicolumn{2}{c|}{INRIA} & \multicolumn{2}{c}{LEVIR-CD}
\\ 
 &  &  & IoU & F1 & IoU & F1
\\
\midrule
& & & 81.31 & 89.62 & 84.82 & 91.76
\\
\checkmark & & & 81.78 & 89.91 & 85.37 & 92.09
\\
\checkmark & \checkmark & & 82.05 & 90.13 & 85.73 & 92.32
\\
\checkmark & \checkmark & \checkmark & \textbf{82.59} & \textbf{90.46} & \textbf{86.19} & \textbf{92.59}
\\
\bottomrule
\end{tabular}
}
\end{table}

%% file: table/table_training_data.tex
\begin{table}[!tbhp]
\centering
\caption{The effects of different pre-training datasets. Performances are validated on the WHU and INRIA test sets. BX means building extraction data, while CD means change detection data.
}
\label{tab:data}
\resizebox{0.8\linewidth}{!}{
\begin{tabular}{c|cc|cc}
\toprule
\multirow{2}{*}{Data} & \multicolumn{2}{c|}{INRIA} & \multicolumn{2}{c}{LEVIR-CD}
\\ 
 & IoU & F1 & IoU & F1
\\
\midrule
- & 81.87 & 89.99 & 85.33 & 92.08
\\
BX & 82.40 & 90.33 & 85.70 & 92.30
\\
CD & 82.19 & 90.21 & 85.97 & 92.46
\\
BX \& CD  & \textbf{82.59} & \textbf{90.46} & \textbf{86.19} & \textbf{92.59}
\\
\bottomrule
\end{tabular}
}
\end{table}

%% file: table/table_generalization.tex
\begin{table}[t]
\centering
\caption{
The generalization performance on the WHU-EAST ASIA (building extraction) and SECOND (change detection) datasets, both of which have been entirely excluded from the training set. For comparative purposes, we have included some state-of-the-art methods that employ fully supervised training. The first section of the table presents the performance for building extraction, while the second delineates the performance for change detection.
}
\label{tab:table_generalization}
\resizebox{0.9\linewidth}{!}{
    \begin{tabular}{c|cccc}
    \toprule
    Method        & P & R & F1    & IoU
    \\ 
    \midrule
    UNet & 84.73 & 80.28    & 82.44   & 70.13 \\
    Segformer       & 84.33 & 82.75    & 83.53   & 71.72 \\
    Deeplab-v3       & 85.21 & 82.65    & 83.91   & 72.28 \\
    Buildformer     & 85.98 & 82.14    & 84.01   & 72.44 \\
    \midrule
    Zero-shot       & 87.69 & 79.35   & 83.31   & 71.42 \\
    Linear Probes    &85.69 &82.85 &84.25 &72.78 \\
    Fine-tuning     & \textbf{86.09} &\textbf{83.96} &\textbf{85.01} & \textbf{73.93}   \\ 
    \midrule
    \midrule
    BiT             & 80.42 & 73.42    & 76.76   & 62.28 \\
    TinyCD          & 79.38 & 71.71    & 75.35   & 60.45 \\
    Changeformer    & 81.71 & 74.85    & 78.13   & 64.11 \\
    Changer-Ex      & 81.80 & 76.22    & 78.90   & 65.17 \\ 
    \midrule
    Zero-shot           & 81.61 & 67.44    & 73.91   & 58.96 \\
    Linear Probes  &82.78 &78.36 &80.51 &67.38 \\
    Fine-tuning &\textbf{84.83} &\textbf{79.79} &\textbf{82.23} &\textbf{69.83}  \\
    \bottomrule
    \end{tabular}
}
\end{table}

%% file: secs/06-conclusion.tex
\section{Conclusion}

In this paper, we introduce a general building interpretation model, RSBuilding, developed from the perspective of the foundation model, 
which realizes the tasks of building extraction and building change detection within a unified framework. 
Our objective is to move beyond meticulously designed branches or modules, and instead utilize a unified foundation model to enhance cross-scene generalization and task universality during building understanding.
Specifically, we adhere to the mainstream Transformer architecture by employing existing visual foundation models' encoders with general knowledge to extract image features. We have designed a straightforward multi-level feature sampler to enrich scale information.
To unify task representation, we propose a cross-attention decoder based on task prompts. 
Regarding training data and strategies for the foundation model, we have addressed the current issue of limited datasets containing annotations for both tasks simultaneously, by developing a federated training strategy that enables the model to converge smoothly. This training strategy can enhance the complementarity between the two interpretation tasks.
Experimental results demonstrate that RSBuilding can achieve state-of-the-art performance on multiple diverse source datasets and exhibits robust zero-shot generalization ability.